# Next Generation Reservoir Computing

Daniel J. Gauthier,[1,2,*] Erik Bollt,[3,4] Aaron Griffith,[1] and Wendson A.S. Barbosa[1]

[1]The Ohio State University, Department of Physics, 191 West Woodruff Ave., Columbus, OH 43210, USA.

[2]ResCon Technologies, LLC, PO Box 21229, Columbus, OH 43221, USA

[3]Clarkson University, Department of Electrical and Computer Engineering, Potsdam, NY 13669

[4]Clarkson Center for Complex Systems Science (C$^3$S$^2$), Potsdam, NY 13699, USA

*Correspondence to: gauthier.51@osu.edu

July 21, 2021

**Abstract:**

Reservoir computing is a best-in-class machine learning algorithm for processing information generated by dynamical systems using observed time-series data. Importantly, it requires very small training data sets, uses linear optimization, and thus requires minimal computing resources. However, the algorithm uses randomly sampled matrices to define the underlying recurrent neural network and has a multitude of metaparameters that must be optimized. Recent results demonstrate the equivalence of reservoir computing to nonlinear vector autoregression, which requires no random matrices, fewer metaparameters, and provides interpretable results. Here, we demonstrate that nonlinear vector autoregression excels at reservoir computing benchmark tasks and requires even shorter training data sets and training time, heralding the next generation of reservoir computing.

## I. Introduction

A dynamical system evolves in time, with examples including the Earth's weather system and human-built devices such as unmanned aerial vehicles. One practical goal is to develop models for forecasting their behavior. Recent machine learning (ML) approaches can generate a model using only observed data, but many of these algorithms tend to be "data hungry," requiring long observation times and substantial computational resources.

Reservoir computing[1,2] is a ML paradigm that is especially well-suited for learning dynamical systems. Even when systems display chaotic[3] or complex spatio-temporal behaviors[4], which are considered the hardest-of-the-hard problems, an optimized reservoir computer (RC) can handle them with ease.

As described in greater detail in the next section, an RC is based on a recurrent artificial neural network with a pool of interconnected neurons – the reservoir, an input layer feeding observed data **X** to the network, and an output layer weighting the network states as shown in Fig. 1. To avoid the vanishing gradient problem[5] during training, the RC paradigm randomly assigns the input-layer and reservoir link weights. Only the weights of the output links $\mathbf{W}_{out}$ are trained via a regularized linear least-squares optimization procedure[6]. Importantly, the regularization parameter $\alpha$ is set to prevent overfitting to the training data in a controlled and well understood manner and makes the procedure noise tolerant. RCs perform as well as other ML methods, such as Deep Learning, on dynamical systems tasks but have substantially smaller data set requirements and faster training times[7,8].

Using random matrices in an RC presents problems: many perform well, but others not all and there is little guidance to select good or bad matrices. Furthermore, there are several RC metaparameters that can greatly affect its performance and require optimization[9]. Recent work suggests that good matrices and metaparameters can be identified by determining whether the reservoir dynamics $r$ synchronizes in a generalized sense to **X** [10,11], but there are no known design rules for obtaining generalized synchronization.

Recent RC research has identified requirements for realizing a general, universal approximator of dynamical systems. An universal approximator can be realized using an RC with nonlinear activation at nodes in the recurrent network and an output layer (known as the feature vector) that is a weighted linear sum of the network nodes under the weak assumptions that the dynamical system has bounded orbits[12].

Less appreciated is the fact that an RC with *linear* activation nodes combined with a feature vector that is a weighted sum of *nonlinear* functions of the reservoir node values is an equivalently powerful universal approximator[12,13]. Furthermore, such an RC is mathematically identical to a nonlinear vector autoregression (NVAR) machine[14]. Here, no reservoir is required: the feature vector of the NVAR consists of $k$ time-delay observations of the dynamical system to be learned and nonlinear functions of these observations, as illustrated in Fig. 1, a surprising result given the apparent lack of a reservoir!

These results are in the form of an existence proof: There exists an NVAR that can perform equally well as an optimized RC and, in turn, the RC is implicit in a NVAR. Here, we demonstrate that it is easy to design a well-performing NVAR for three challenging RC benchmark problems: 1) forecasting the short-term dynamics; 2) reproducing the long-term 'climate' of a chaotic system (that is, reconstructing the attractors shown in Fig. 1); and 3) inferring the behavior of unseen data of a dynamical system.

Predominantly, the recent literature has focused on the first benchmark of short-term forecasting of stochastic processes time-series data[12], but the importance of high-accuracy forecasting and inference of unseen data cannot be overstated. The NVAR, which we call the Next Generation RC (NG-RC), displays state-of-the-art performance on these tasks because it is associated with an implicit RC, and uses exceedingly small data sets and side-steps the random and parametric difficulties of directly implementing a traditional RC.

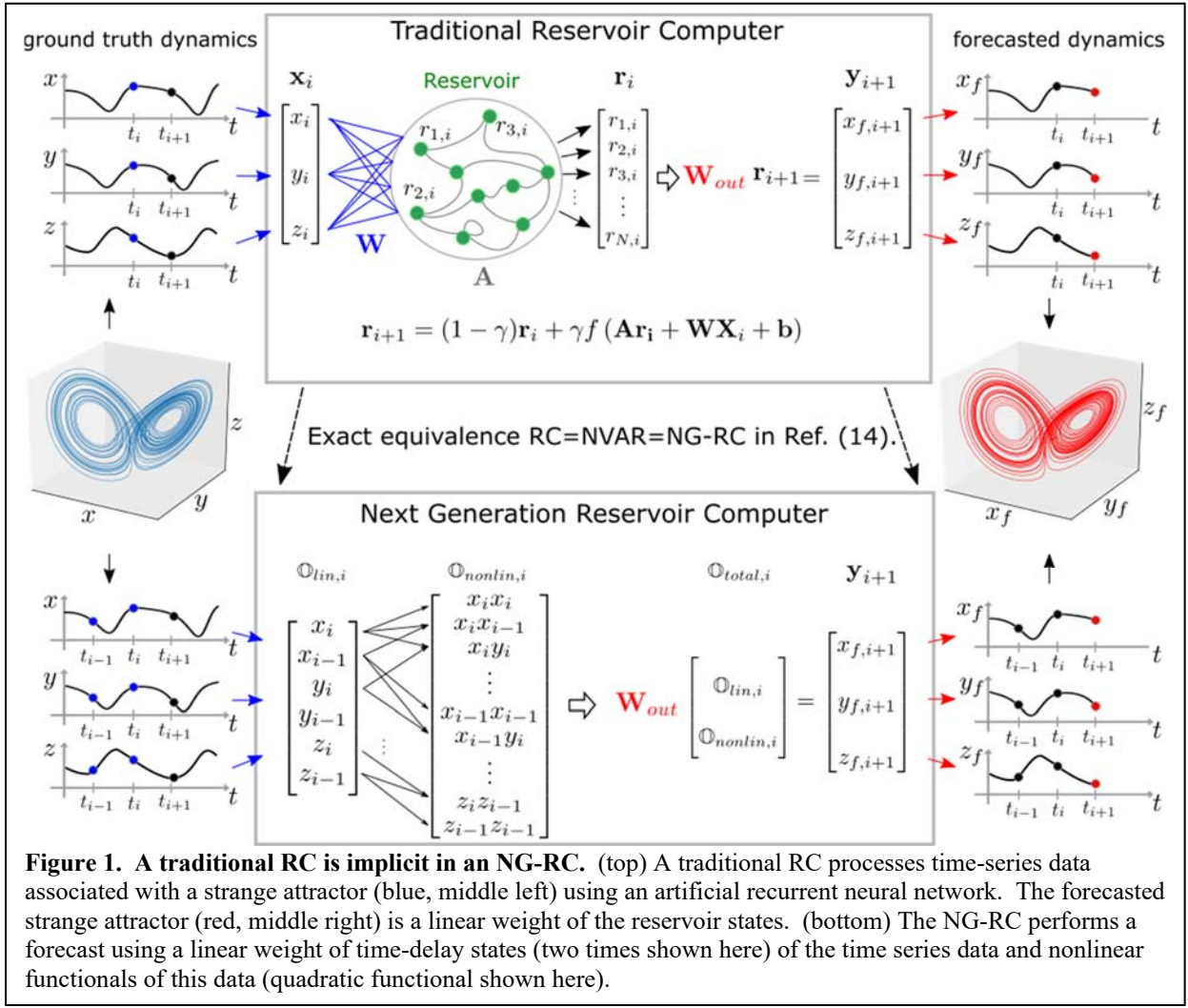

**Figure 1. A traditional RC is implicit in an NG-RC.** (top) A traditional RC processes time-series data associated with a strange attractor (blue, middle left) using an artificial recurrent neural network. The forecasted strange attractor (red, middle right) is a linear weight of the reservoir states. (bottom) The NG-RC performs a forecast using a linear weight of time-delay states (two times shown here) of the time series data and nonlinear functionals of this data (quadratic functional shown here).

In the next section, we briefly review traditional RCs and introduce an RC with linear reservoir nodes and a nonlinear output layer. In Sec. III, we introduce the NG-RC and discuss the remaining metaparameters. We then go on to introduce two model systems we use to showcase the performance of the NG-RC (Sec. IV) and present our findings (Sec. V). Finally, we discuss the implications of our work and future directions in Sec. VI.

## II. Background: Reservoir Computing

The purpose of an RC illustrated in the top panel of Fig. 1 is to broadcast input data **X** into the higher-dimensional reservoir network composed of $N$ interconnected nodes, and then to combine the resulting reservoir state into an output **Y** that closely matches a desired output $\mathbf{Y}_d$. The strength of the node-to-node connections, represented by the connectivity (or adjacency) matrix **A**, are chosen randomly and kept fixed. The data to be processed **X** is broadcast into the reservoir through the input layer with fixed random coefficients **W**. The reservoir is a dynamical system whose dynamics can be represented by

$$\mathbf{r}_{i+1} = (1-\gamma)\mathbf{r}_i + \gamma f(\mathbf{A}\mathbf{r}_i + \mathbf{W}\mathbf{X}_i + \mathbf{b}), \tag{1}$$

where $\mathbf{r}_i = [r_{1,i}, r_{2,i}, \ldots, r_{N,i}]^T$ is an N-dimensional vector with component $r_{j,i}$ representing the state of the $j^{th}$ node at the time $t_i$, $\gamma$ is the decay rate of the nodes, $f$ an activation function applied to each vector component, and **b** is a node bias vector. For simplicity, we choose $\gamma$ and **b** the same for all nodes. Here, time is discretized at a finite sample time $dt$ and $i$ indicates the $i^{th}$ time step, so that $dt = t_{i+1} - t_i$. Thus, the notations $\mathbf{r}_i$ and $\mathbf{r}_{i+1}$ represent the reservoir state in consecutive time steps. The reservoir can also equally well be represented by continuous-time ordinary differential equations that may include the possibility of delays along the network links[15].

The output layer expresses the RC output $\mathbf{Y}_{i+1}$ as a linear transformation of a feature vector $\mathbb{O}_{total,i+1}$, constructed from the reservoir state $\mathbf{r}_{i+1}$, through the relation

$$\mathbf{Y}_{i+1} = \mathbf{W}_{out}\mathbb{O}_{total,i+1}, \tag{2}$$

where $\mathbf{W}_{out}$ is the output weight matrix and the subscript "*total*" indicates that it can be composed of constant, linear and nonlinear terms as explained below. The standard approach, commonly used in the RC community, is to choose a nonlinear activation function such as $f(x) = \tanh(x)$ for the nodes and a linear feature vector $\mathbb{O}_{total,i+1} = \mathbb{O}_{lin,i+1} = \mathbf{r}_{i+1}$ in the output layer. The RC is trained using supervised training via regularized least-square regression. Here, the training data points generate a block of data contained in $\mathbb{O}_{total}$ and we match **Y** to the desired output $\mathbf{Y}_d$ in a least-square sense using Tihkonov regularization so that $\mathbf{W}_{out}$ is given by

$$\mathbf{W}_{out} = \mathbf{Y}_d \mathbb{O}_{total}^T (\mathbb{O}_{total}\mathbb{O}_{total}^T + \alpha \mathbf{I})^{-1}, \tag{3}$$

where the regularization parameter $\alpha$, also known as ridge parameter, is set to prevent overfitting to the training data and **I** is the identity matrix.

*Linear reservoir + Nonlinear output*

A different approach to RC is to move the nonlinearity from the reservoir to the output layer[12,14]. This approach is an equivalently powerful universal approximator. In this case, the reservoir nodes are chosen to have a linear activation function $f(\mathbf{r}) = \mathbf{r}$, while the feature vector $\mathbb{O}_{total}$ becomes nonlinear. A simple example of such RC is to extend the standard linear feature vector to include the squared values of all nodes, which are obtained through the Hadamard product $\mathbf{r} \odot \mathbf{r} = [r_1^2, r_2^2, \ldots, r_N^2]^T$ [14]. Thus, the nonlinear feature vector is given by

$$\mathbb{O}_{total} = \mathbf{r} \oplus (\mathbf{r} \odot \mathbf{r}) = [r_1, r_2, \ldots, r_N, r_1^2, r_2^2, \ldots, r_N^2]^T, \tag{4}$$

where $\oplus$ represents the vector concatenation operation. A linear reservoir with a nonlinear output is an equivalently powerful universal approximator[12] and shows comparable performance to the standard RC[14].

## III. The NG-RC

The NVAR, which we call a next-generation reservoir computer (NG-RC), creates a feature vector directly from the discretely sample input data with no need of a neural network. Here, $\mathbb{O}_{total} = c \oplus \mathbb{O}_{lin} \oplus \mathbb{O}_{nonlin}$, where $c$ is a constant and $\mathbb{O}_{nonlin}$ is a nonlinear part of the feature vector. Like a traditional RC, the output at $t_i$ is obtained by training the features given using Eq. (2) and Tihkonov regularization. We now discuss forming these features.

*Linear features*

The linear features $\mathbb{O}_{lin,i}$ at time step $i$ is composed of observations of the input vector **X** at the current and at $k$-1 previous times steps spaced by $s$, where ($s$-1) is the number of skipped steps between consecutive observations. If $\mathbf{X}_i = [x_{1,i}, x_{2,i}, \ldots, x_{d,i}]^T$ is a $d$-dimensional vector, $\mathbb{O}_{lin,i}$ has $d\,k$ components and is given by

$$\mathbb{O}_{lin,i} = \mathbf{X}_i \oplus \mathbf{X}_{i-s} \oplus \mathbf{X}_{i-2s} \oplus \ldots \oplus \mathbf{X}_{i-(k-1)s}. \tag{5}$$

Based on the general theory of universal approximators[12,16], $k$ should be taken to be infinitely large. However, it is found in practice that the Volterra series converges rapidly and hence truncating $k$ to small values does not incur large error. This can also be motivated by considering numerical integration methods of ordinary differential equations where only a few sub-intervals (steps) in a multi-step integrator are needed to obtain high accuracy. We do not sub-divide the step size here, but this analogy motivates why small values of $k$ might give good performance in the forecasting tasks considered below.

An important aspect of the NG-RC is that its "warm-up" period only contains $s\,k$ time steps, which are needed to create the feature vector for the first point to be processed. This is a dramatically shorter warm-up period in comparison to traditional RCs, where longer warm up times are needed to ensure that the reservoir state does not depends on the RC initial conditions. For example, with $s$=1 and $k$=2 as used for some examples below, only 2 warm up data points are needed. A typical warm up time in traditional RC for the same task can be upwards of $10^3$ to $10^5$ data points[9,10]. A reduced warmup time is especially important in situations where it is difficult to obtain data or collecting additional data is too time consuming.

For the case of a driven dynamical system, $\mathbb{O}_{lin}(t)$ also includes the drive signal[17]. Similarly, for a system in which one or more accessible system parameters are adjusted, $\mathbb{O}_{lin}(t)$ also includes these parameters[17,18].

*Nonlinear features*

The nonlinear part $\mathbb{O}_{nonlin}$ of the feature vector is a nonlinear functional of $\mathbb{O}_{lin}$. While there is great flexibility in choosing the nonlinear functionals, we find that polynomials provide good prediction ability. Polynomial functionals are the basis of a Volterra representation for dynamical systems[16] and hence they are a natural starting point. We find that low-order polynomials are enough to obtain high performance.

All monomials of the quadratic polynomial, for example, are captured by the outer product $\mathbb{O}_{lin} \otimes \mathbb{O}_{lin}$, which is a symmetric matrix with $(dk)^2$ elements. A quadratic nonlinear feature vector $\mathbb{O}_{nonlinear}^{(2)}$, for example, is composed of the $(dk)(dk+1)/2$ unique monomials of $\mathbb{O}_{lin} \otimes \mathbb{O}_{lin}$, which are given by the upper triangular elements of the outer product tensor. We define $[\otimes]$ as the operator that collects the unique monomials in a vector. Using this notation, a $p$-order polynomial feature vector is given by

$$\mathbb{O}^{(p)}_{nonlinear} = \mathbb{O}_{lin}[\otimes]\mathbb{O}_{lin}[\otimes] \ldots [\otimes]\mathbb{O}_{lin} \quad \text{with } \mathbb{O}_{lin} \text{ appearing } p \text{ times.} \tag{6}$$

Recently, it was mathematically proven that the NVAR method is equivalent to a linear RC with polynomial nonlinear readout[14]. This means that every NVAR implicitly defines the connectivity matrix and other parameters of a traditional RC described in Sec. II, and that every linear polynomial-readout RC can be expressed as an NVAR. However, the traditional RC is more computationally expensive and requires optimizing many meta-parameters, while the NG-RC is more efficient and straightforward. The NG-RC is doing the same work as the equivalent traditional RC with a full recurrent neural network, but we do not need to find that network explicitly or do any of the costly computation associated with it.

## IV. Model Systems and Tasks

*Lorenz63*

For one of the forecasting tasks and the inference task discussed in the next section, we generate training and testing data by numerically integrating a simplified model of a weather system[19] developed by Lorenz in 1963. It consists of a set of three coupled nonlinear differential equations given by

$$\begin{aligned} \dot{x} &= 10(y-x), \\ \dot{y} &= x(28-z) - y, \\ \dot{z} &= xy - 8z/3, \end{aligned} \tag{7}$$

where the state $\mathbf{X}(t) \equiv [x(t), y(t), z(t)]^T$ is a vector whose components are Rayleigh-Bénard convection observables. It displays deterministic chaos, sensitive dependence to initial conditions - the so-called 'butterfly effect' - and the phase space trajectory forms a strange attractor shown in Fig. 1.

For future reference, the Lyapunov time for Eq. (7), which characterizes the divergence timescale for a chaotic system, is 1.1 time units. Below, we refer to this system as Lorenz63.

*Double-scroll electronic circuit*

We also explore using the NG-RC to predict the dynamics of a double-scroll electronic circuit[20] whose behavior is governed by

$$\begin{aligned} \dot{V}_1 &= V_1/R_1 - \Delta V/R_2 - 2I_r\sinh(\alpha\Delta V), \\ \dot{V}_2 &= \Delta V/R_2 + 2I_r\sinh(\alpha\Delta V) - I, \\ \dot{I} &= V_2 - R_4 I, \end{aligned} \tag{8}$$

in dimensionless form, where $\Delta V = V_1 - V_2$. Here, we use the parameters $R_1 = 1.2$, $R_2 = 3.44$, $R_4 = 0.193$, $\alpha = 11.6$, and $I_r = 2.25 \times 10^{-5}$, which give a Lyapunov time 7.81 time units.

We select this system because the vector field is not of a polynomial form and $\Delta V$ is large enough at some times that a truncated Taylor series expansion of the sinh function gives rise to large differences in the predicted attractor. This task demonstrates that the polynomial form of the feature vector works for nonpolynomial vector fields as expected from the theory of Volterra representations of dynamical systems[16].

*Forecasting task*

In the two forecasting tasks presented below, we use a NG-RC to forecast the dynamics of Lorenz63 and the double-scroll system using one-step-ahead prediction. We start with a "listening" phase, seeking a solution to $\mathbf{X}(t+dt) = \mathbf{W}_{out}\mathbb{O}_{total}(t)$, where $\mathbf{W}_{out}$ is found using Tihkonov regularization[6]. During the "forecasting" or "testing" phase, the components of $\mathbf{X}(t)$ are no longer provided to the NG-RC and the predicted output is fed back to the input. Now, the NG-RC is an autonomous dynamical system that predicts the systems' dynamics if training is successful.

The total feature vector used for the Lorenz63 forecasting task is given by

$$\mathbb{O}_{total} = c \oplus \mathbb{O}_{lin} \oplus \mathbb{O}^{(2)}_{nonlinear}, \qquad \text{Lorenz63} \qquad (9)$$

which has [1+ *d k*+(*d k*) (*d k*+1)/2] components. For the double-scroll system forecasting task, we notice that the attractor has odd symmetry and has zero mean for all variables for the parameters we use. To respect these characteristics, we take

$$\mathbb{O}_{total} = \mathbb{O}_{lin} \oplus \mathbb{O}^{(3)}_{nonlinear}, \qquad \text{double-scroll} \qquad (10)$$

which has [*d k*+(*d k*) (*d k*+1) (*d k*+2)/6] components.

For these forecasting tasks, the NG-RC learns simultaneously the vector field and an efficient one-step-ahead integrator to find a mapping from one time to the next without having to learn each separately as in other nonlinear state estimation approaches[21–24]. The one-step-ahead mapping is known as the *flow* of the dynamical system and hence the NG-RC learns the flow. To allow the NG-RC to focus on the subtle details of this process, we use a simple Euler-like integration step as a lowest-order approximation to a forecasting step by modifying Eq. (2) so that the NG-RC learns the difference between the current and future step. To this end, Eq. (2) is replaced by

$$\mathbf{Y}_{i+1} = \mathbf{Y}_i + \mathbf{W}_{out}\mathbb{O}_{total,i+1}. \qquad (11)$$

*Inference task*

In the third task, we provide the NG-RC with all three Lorenz63 variables during training with the goal of inferring one of the variables from the others. During testing, we only provide it with the *x* and *y* variables and infer the *z* variable. This task is important for applications where it is possible to obtain high-quality information about a dynamical variable in a laboratory setting, but not in a field deployment. In the field, the observable sensory information is used to infer the missing data.

## V. Results

*Task 1: Forecasting Lorenz63*

For the first task, the ground-truth Lorenz63 strange attractor shown in Fig. 2a. The training phase uses only the data shown in Figs. 2b-d, which consists of 400 data points for each variable with $dt$=0.025, $k$=2, and $s$=1. The training compute time is <10 ms using Python running on a single-core desktop processor (see Methods). Here, $\mathbb{O}_{total}$ has 28 components and $\mathbf{W}_{out}$ has dimension (3×28). The set needs to be long enough for the phase-space trajectory to explore both 'wings' of the strange attractor. The plot is overlayed with the NG-RC predictions during training; no difference is visible on this scale.

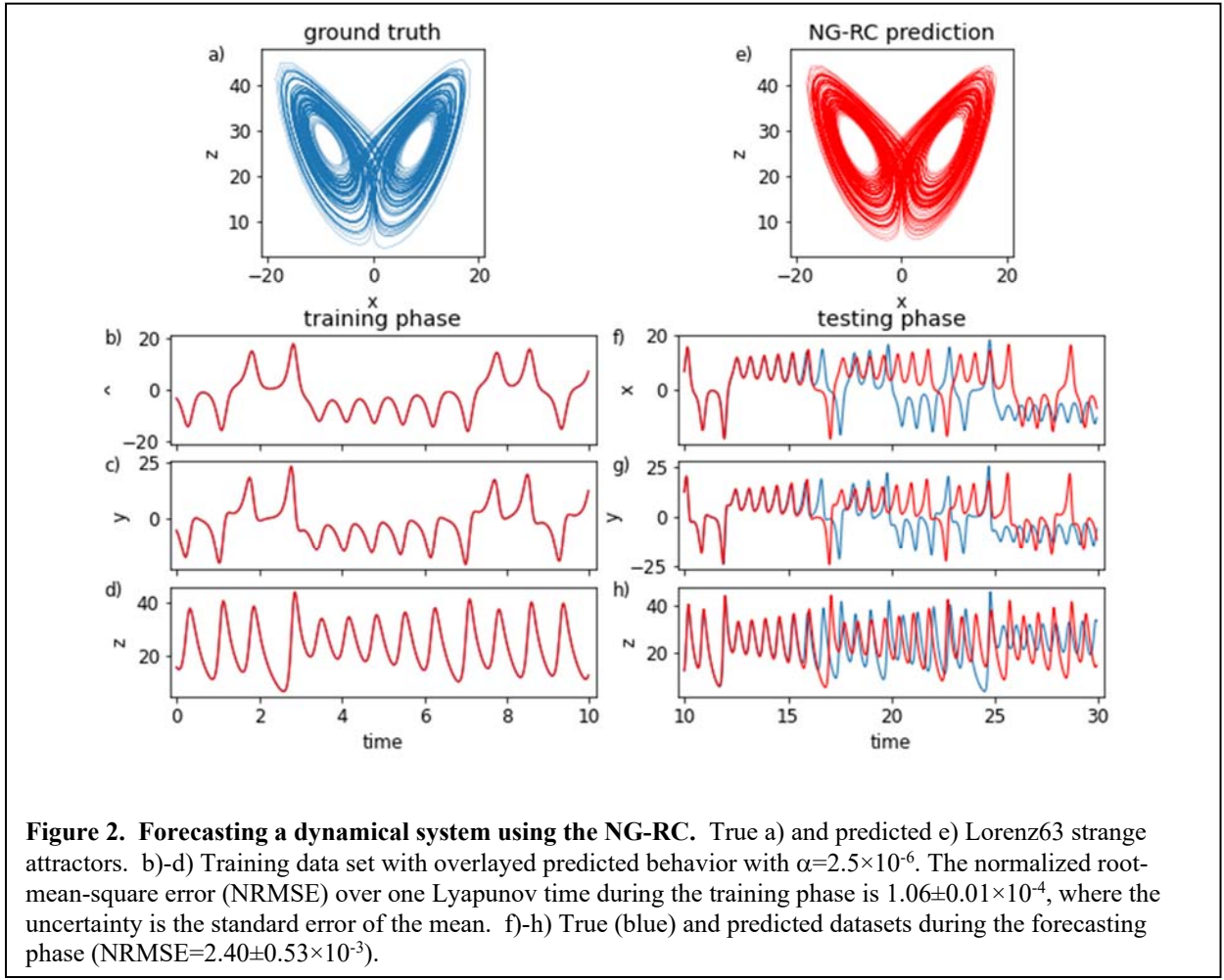

**Figure 2. Forecasting a dynamical system using the NG-RC.** True a) and predicted e) Lorenz63 strange attractors. b)-d) Training data set with overlayed predicted behavior with $\alpha$=2.5×10$^{-6}$. The normalized root-mean-square error (NRMSE) over one Lyapunov time during the training phase is 1.06±0.01×10$^{-4}$, where the uncertainty is the standard error of the mean. f)-h) True (blue) and predicted datasets during the forecasting phase (NRMSE=2.40±0.53×10$^{-3}$).

The NG-RC is then placed in the prediction phase; a qualitative inspection of the predicted (Fig. 2e) and true (Fig. 2a) strange attractors shows that they are very similar, indicating that the NG-RC reproduces the long-term 'climate' of Lorenz63 (benchmark problem 2). As seen in Figs. 2f-h, the NG-RC does a good job of predicting Lorenz63 (benchmark 1), comparable to an optimized traditional RC[3,9,10] with 100's to 1,000's of reservoir nodes. The NG-RC forecasts well out to ~5 Lyapunov times.

In the Supplementary Information, we give other quantitative measurements of the accuracy of the attractor reconstruction. We also give there the values of $\mathbf{W}_{out}$; there are many components that have substantial weights and that do not appear in the vector field of Eq. (7), where the vector field is the right-hand-side of the differential equations. This gives quantitative information regarding the difference between the flow and the vector field.

*Task 2: Forecasting the double-scroll system*

Because the Lyapunov time for the double-scroll system is much longer than for the Lorenz63 system, we extend the training time of the NG-RC from 10 to 100 units to keep the number of Lyapunov times covered during training similar for both cases. To ensure a fair comparison to the Lorenz63 task, we set $dt = 0.25$. With these two changes, and the use of the cubic monomials, as given in Eq. 10, with $d=3$, $k=2$, and $s=1$ for a total of 62 features in $\mathbb{O}_{total}$, the NG-RC uses 400 data points for each variable during training, exactly as in the Lorenz63 task.

Other than these modifications, our method for using the NG-RC to forecast the dynamics of this system proceed exactly as for the Lorenz63 system. The results of this task are displayed in Fig. 3, where it is seen that the NG-RC shows similar predictive ability on the double-scroll system as in the Lorenz63 system, where other quantitative measures of accurate attractor reconstruction is given in the Supplementary Information as well as the components of $\mathbf{W}_{out}$.

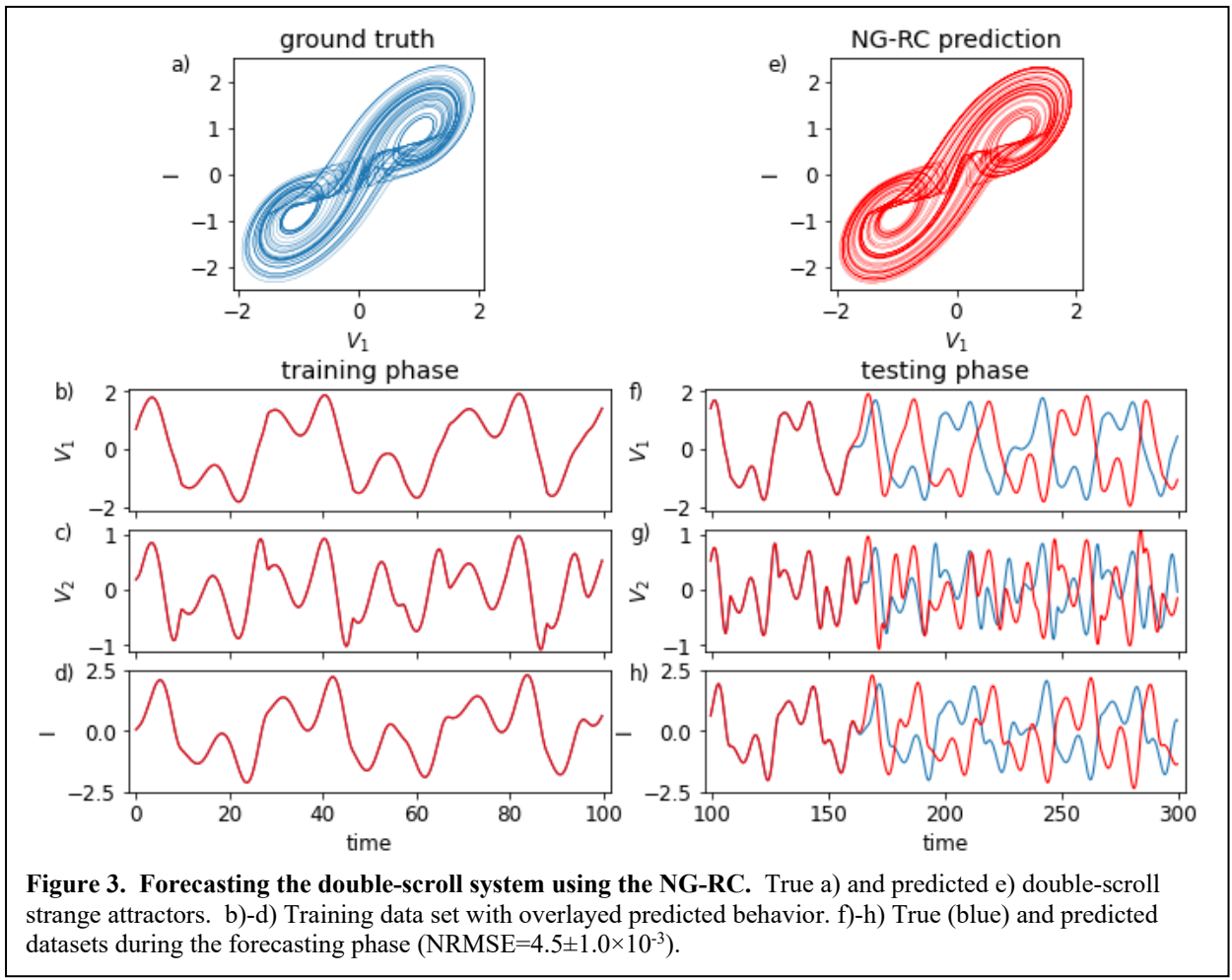

**Figure 3. Forecasting the double-scroll system using the NG-RC.** True a) and predicted e) double-scroll strange attractors. b)-d) Training data set with overlayed predicted behavior. f)-h) True (blue) and predicted datasets during the forecasting phase (NRMSE=$4.5\pm1.0\times10^{-3}$).

*Task 3: Inferring unseen Lorenz63 dynamics*

In the last task, we infer dynamics not seen by the NG-RC during the testing phase. Here, we use $k=4$ and $s=5$ with $dt=0.05$ to generate an embedding of the full attractor to infer the other component, as informed by Takens' embedding theorem[25]. We provide the $x$, $y$, and $z$ variables during training and we again observe that a short training data set of only 400 points is enough to obtain good performance as shown in Fig. 4c, where the training data set is overlayed with the NG-RC predictions. Here, the total feature vector has 45 components and hence $\mathbf{W}_{out}$ has dimension ($1\times45$). During testing phase, we only provide the NG-RC with the $x$ and $y$ components (Figs. 4d and e) and predict the $z$ component (Fig. 4f). The performance is nearly identical during the testing phase. The components of $\mathbf{W}_{out}$ for this task are given in the Supplementary Information.

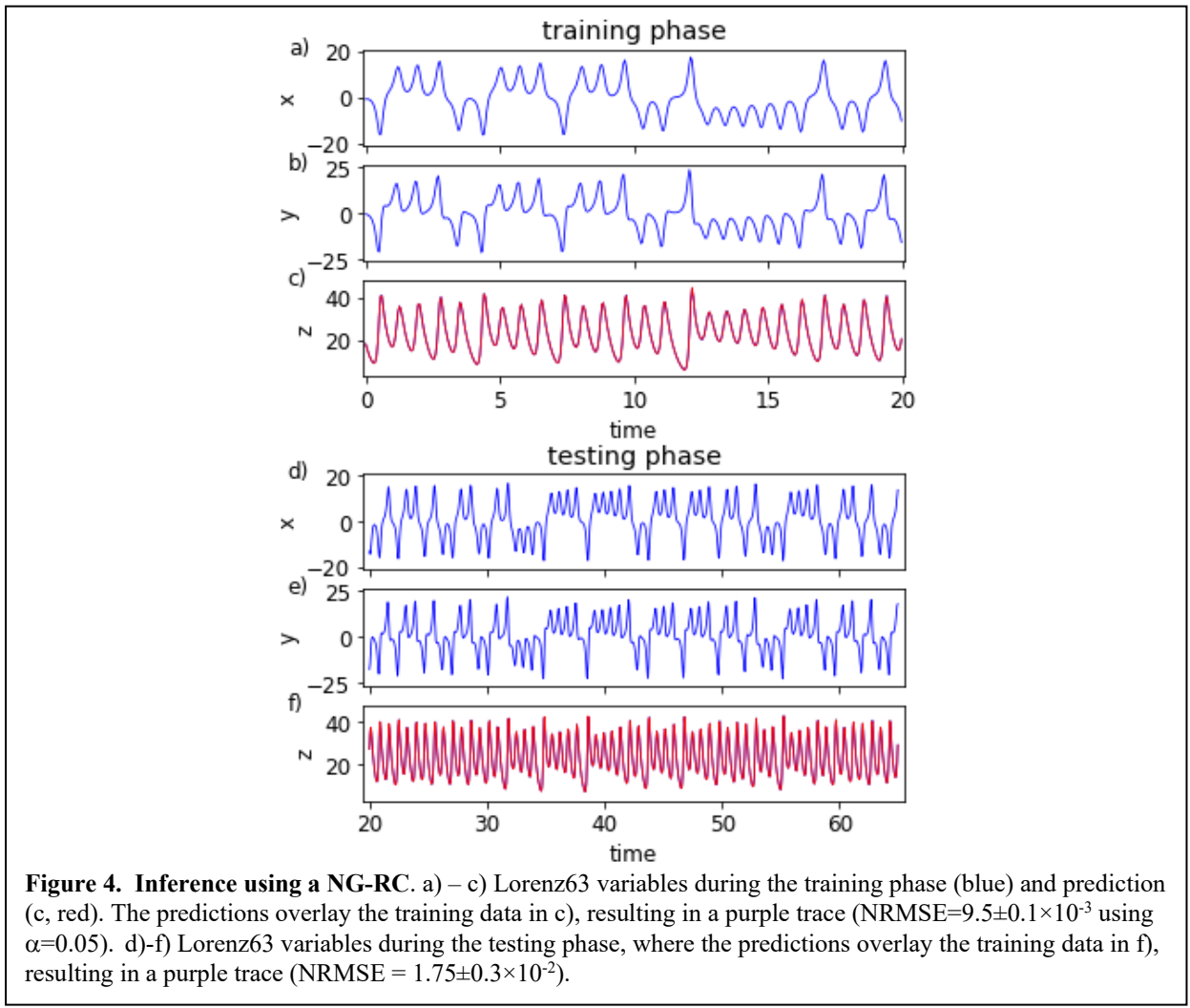

**Figure 4. Inference using a NG-RC.** a) – c) Lorenz63 variables during the training phase (blue) and prediction (c, red). The predictions overlay the training data in c), resulting in a purple trace (NRMSE=$9.5\pm0.1\times10^{-3}$ using $\alpha=0.05$). d)-f) Lorenz63 variables during the testing phase, where the predictions overlay the training data in f), resulting in a purple trace (NRMSE = $1.75\pm0.3\times10^{-2}$).

## VI. Discussion

The NG-RC is computationally faster than a traditional RC because the feature vector size is much smaller, meaning there are fewer adjustable parameters that must be determined as discussed in the Supplementary Materials. We believe that the training data set size is reduced precisely because there are fewer fit parameters. Also, as mentioned above, the warmup and training time are shorter, thus reducing the computational time. Finally, the NG-RC has fewer metaparameters to optimize, thus avoiding the computational costly optimization procedure in a high-dimensional parameter space. As detailed in the Supplemental Materials, we estimate the computational complexity for the Lorenz63 forecasting task and find that the NG-RC is approximately 33-162 times less costly to simulate than a typical already efficient traditional RC[9], and over $10^6$ times less costly for a high-accuracy traditional RC[10] for a single set of metaparameters. For the double-scroll system, where the NG-RC has a cubic nonlinearity, and hence more features, the improvement is a more modest factor of 8-41 than a typical efficient traditional RC[9] for a single set of metaparameters.

The NG-RC builds on previous work on nonlinear system identification. It is most closely related to multi-input, multiple-output nonlinear autoregression with exogenous inputs (NARX) studied since the 1980's[17]. A crucial distinction is that Tikhonov regularization is not used in the NARX approach and there is no theoretical underpinning of a NARX to an implicit RC. Our NG-RC fuses the best of the NARX methods with modern regression methods, which is needed to obtain the good performance demonstrated here. We mention that Pyle *et al.*[26] recently found good performance with a simplified NG-RC but without the theoretical framework and justification presented here.

In other related work, there has been a revival of research on data driven linearization methods[27] that represent the vector field by projecting onto a finite linear subspace spanned by simple functions, usually monomials. Notably, Ref. [21] uses least-square while recent work uses LASSO[22,23] or information theoretic methods[28] to simplify the model. The goal of these methods is to model the vector field from data, as opposed to the NG-RC developed here that forecasts over finite time steps and thus learns the flow of the dynamical system. In fact, some of the large-probability components of $\mathbf{W}_{out}$ (Supplementary Information) can be motivated by the terms in the vector-field but many others are important demonstrating that the NG-RC-learned flow is different from the vector field.

Some of the components of $\mathbf{W}_{out}$ are quite small, suggesting that several features can be removed using various methods without hurting the testing error. In the NARX literature[17], it is suggested that a practitioner start with the lowest number of terms in the feature vector and add terms one-by-one, keeping only those terms that reduce substantially the testing error based on an arbitrary cutoff in the observed error reduction. This procedure is tedious and ignores possible correlations in the components. Other theoretically justified approaches include using the LASSO or information theoretic methods mentioned above. The other approach to reducing the size of the feature space is to use the "kernel trick" that is the core of ML via support vector machines[16]. This approach will only give a computational advantage when the dimension of $\mathbb{O}_{total}$ is much greater than the number of training data points, which is not the case in our studies here but maybe relevant in other situations. We will explore these approaches in future research.

Our study only considers data generated by noise-free numerical simulations of models. It is precisely the use of regularized regression that makes this approach noise tolerant: it identifies a model that is a best estimator of the underlying dynamics even with noise or uncertainty. We give results for forecasting the Lorenz63 system when it is strongly driven by noise in the Supplementary Materials, where we observe that the NG-RC learns the equivalent noise-free system as long as α is increased demonstrating the importance of regularization.

We also only consider low-dimensional dynamical systems, but previous work forecasting complex high-dimensional spatial-temporal dynamics[4,7] using a traditional RC suggests that an

NG-RC will excel at this task because of the implicit traditional RC but using smaller datasets and requiring optimizing fewer metaparameters. Furthermore, Pyle *et al*.[26] successfully forecast the behavior a multi-scale spatial-temporal system using an approach similar to the NG-RC.

Our work has important implications for learning dynamical systems because there are fewer metaparameters to optimize and the NG-RC only requires extremely short datasets for training. Because the NG-RC has an underlying implicit (hidden) traditional RC, our results generalize to any system for which a standard RC has been applied previously. For example, the NG-RC can be used to create a 'digital twin' for dynamical systems[29] using only observed data or by combining approximate models with observations for data assimilation[30,31]. It can also be used for nonlinear control of dynamical systems[32], which can be quickly adjusted to account for changes in the system, or for speeding up the simulation of turbulence[33].

**Methods**

The exact numerical results presented here, such as unstable steady states (USSs) and NRMSE, will vary slightly depending on the precise software used to calculate them. We calculate the results for this paper using Python 3.7.9, NumPy 1.20.2, and SciPy 1.6.2 on an x86-64 CPU running Windows 10.

**Acknowledgments**

We gratefully acknowledge discussions with Henry Abarbanel, Ingo Fischer, and Kathy Lüdge. *Funding:* DJG is supported by the United States Air Force AFRL/SBRK under Contract No. FA864921P0087. EB is supported by the ARO (N68164-EG) and DARPA. *Author contributions:* DJG optimized the NG-RC, performed the simulations in the main text, and drafted the manuscript. EB conceptualized the connection between an RC and NVAR, helped interpret the data, and edited the manuscript. AG and WSAB helped interpret the data and edited the manuscript. *Competing interests:* DJG is a co-founder of ResCon Technologies, LCC, which is commercializing RCs.

**Code and Data Availability**

All data and code are available on Github (https://github.com/quantinfo/ng-rc-paper-code).

# Supplementary Information for
# Next Generation Reservoir Computing

Daniel J. Gauthier, Erik Bollt, Aaron Griffith, and Wendson A.S. Barbosa

July 21, 2021

**Forecasting Verification**

For the Lorenz63 system, we use two methods for verifying that the forecasted attractor is an accurate representation of the true attractor. The first relies on comparing the unstable steady states of the predicted and true attractors and the second is a qualitative comparison to the return map associated with the strange attractor. For the double-scroll system, we only compare the true and predicted unstable steady states.

*Lorenz63 Unstable Steady States*

The unstable steady states (USSs) of the true Lorenz63 system are determined by setting the derivatives in Eq. (7) to zero and solving for the values of the three variables. There are three solutions given by

$$\mathbf{X}_{uss} = [0,0,0]^T, [\pm\sqrt{\beta(\rho-1)}, \pm\sqrt{\beta(\rho-1)}, (\rho-1)]^T. \quad \text{Lorenz63} \quad (12)$$

We calculate the $L_2$ (Euclidean) distance from the predicted USSs to their corresponding true value. To allow easy comparison of the accuracy of these USSs, we calculate these distances in a uniformly scaled space where Lorenz63 has unit variance. For the model predictions used to generate the attractor shown in Fig. 2, the $L_2$ distance from the zero USS, calculated in a uniformly scaled space where the Lorenz63 system has unit variance, is $1.2 \pm 1.4 \times 10^{-3}$, and the distance between the predicted and true positive (negative) USS is $12.1 \pm 3.1 \times 10^{-4}$ ($7.6 \pm 1.5 \times 10^{-4}$).

*Double-Scroll Unstable Steady States*

We find the USS by setting the derivatives in Eq. (8) to zero and solving for the state variables. This reduces to solving the transcendental equation

$$0 = V_1/R_2 (R_1 - R_4 - R_2) + 2R_1 I_r \sinh(\alpha(1 - R_4/R_1)V_1). \quad (13)$$

For the parameters we use, this yields three solutions, and therefore three USSs given by

$$\mathbf{X}_{uss} = [0, 0, 0]^T, [V_1, \pm V_1 R_4/R_1, \pm V_1/R_1]^T. \quad \text{double-scroll} \quad (14)$$

Because the NG-RC for this task has only odd polynomial powers and no constant term, it is symmetric about the origin and predicts the zero USS exactly. The $L_2$ distance from the true non-zero positive USS, calculated in a uniformly scaled space where the double-scroll system has unit variance, is $2.1 \pm 0.2 \times 10^{-3}$.

*Return map*

The $z$ variable of the Lorenz63 system has a functional relation between successive local maxima. This is demonstrated visually by finding the successive local maxima $M_i$ of $z$, and then plotting $M_i$ with respect to $M_{i+1}$ (*19*). This return map neatly summarizes the long-term behavior of the $z$ variable and comparing two such maps provide a quick way to qualitatively compare two systems, and has been used previously to verify that a trained RC can replicate the Lorenz63 climate (*3*).

The precise shape of this return map depends on the integrator used to find the Lorenz63 solution. In this paper, we use an explicit Runge-Kutta 3(2) integrator for both the Lorenz63 and the double-scroll systems.

To evaluate whether an NG-RC can replicate this long-term behavior, we perform the same procedure on a free-running forecast produced by an NG-RC previously trained on the Lorenz63 system. For both the NG-RC and Lorenz63, we look for maxima in window of 1,000 time units.

The values of the maxima in the discrete-time solutions for both the NG-RC and Lorenz63 depend on the time step $dt$ used for integration, as the true maximum may be achieved in between the discrete time steps. To better reproduce the true Lorenz63 return map and to reduce the effect $dt$ has on the NG-RC return map, we interpolate the $z$ solutions by using a degree-4 spline. The local maxima are then found on this interpolated spline.

Figure 5 shows both return maps. Qualitatively, there is good agreement between the two return maps. The NG-RC return map almost completely obscures the true Lorenz63 return map at this scale. Upon close inspection, we see that the NG-RC return map does not line up precisely with the true Lorenz63 return map. This can be improved by extending the training time of the NG-RC, but difference between the two return maps is already small even when the NG-RC is trained for only 10 time units (400 total data points).

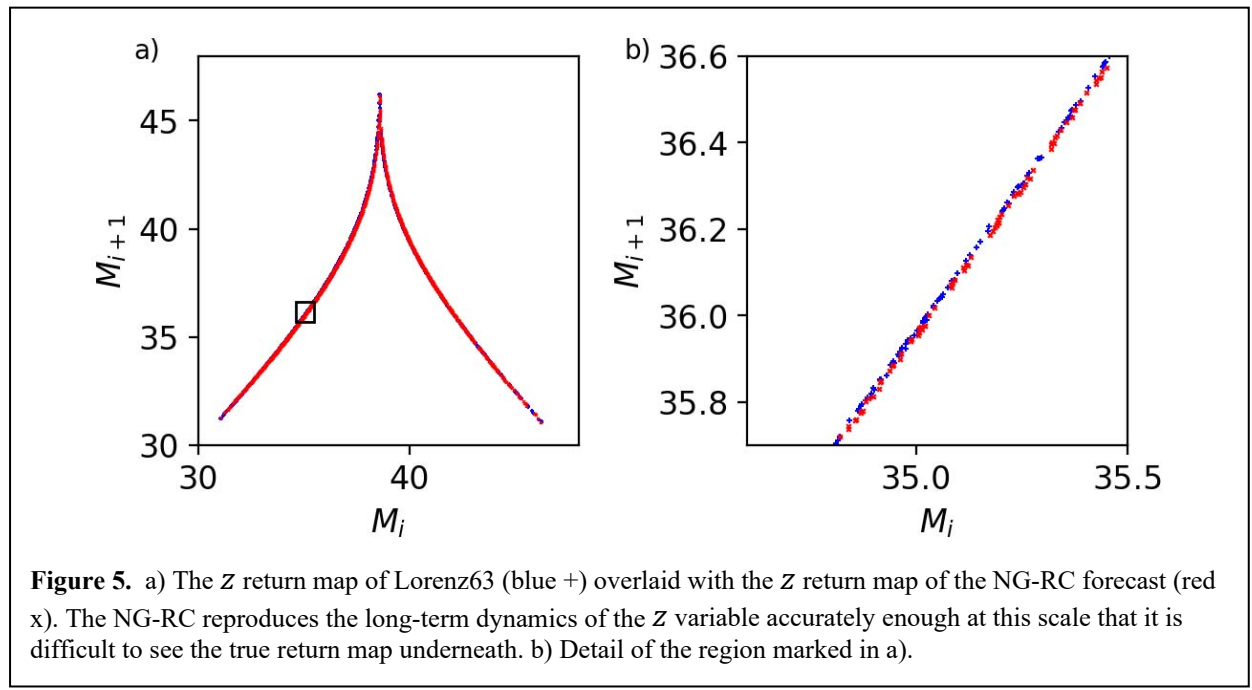

**Figure 5.** a) The $z$ return map of Lorenz63 (blue +) overlaid with the $z$ return map of the NG-RC forecast (red x). The NG-RC reproduces the long-term dynamics of the $z$ variable accurately enough at this scale that it is difficult to see the true return map underneath. b) Detail of the region marked in a).

# Elements of $W_{out}$ for the two tasks

## Forecasting task

Figure 6 shows the *x*, *y*, and *z* components of $\mathbf{W}_{out}$ for the forecasting task presented in Fig. 2 of the main text and Fig. 7 shows a zoom-in of the components. These components vary smoothly with the regularization parameter $\alpha$ over the range we consider in this work. Comparing to the vector field of Eq. 7, there are many substantial components that do not appear directly in the vector field.

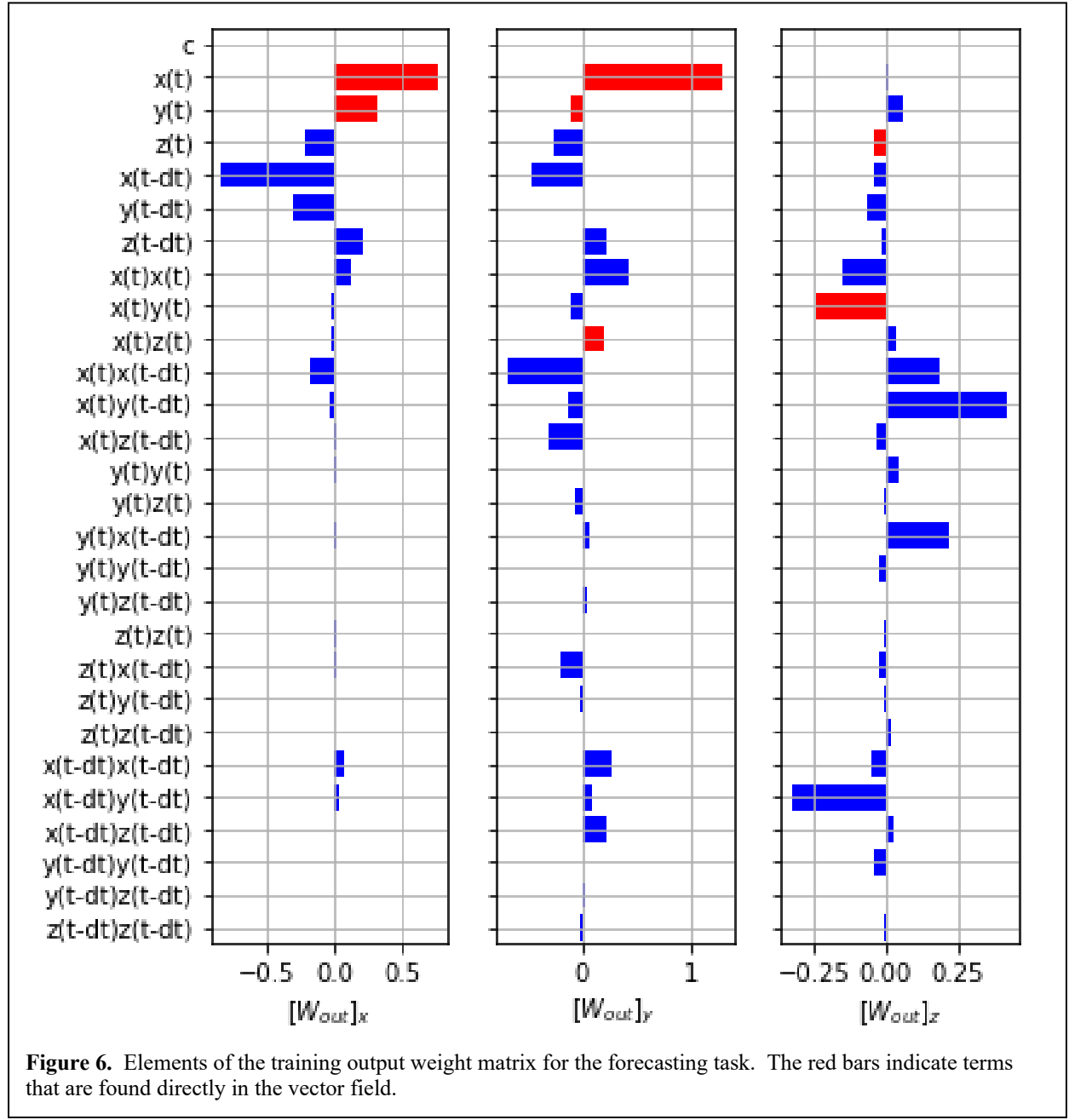

**Figure 6.** Elements of the training output weight matrix for the forecasting task. The red bars indicate terms that are found directly in the vector field.

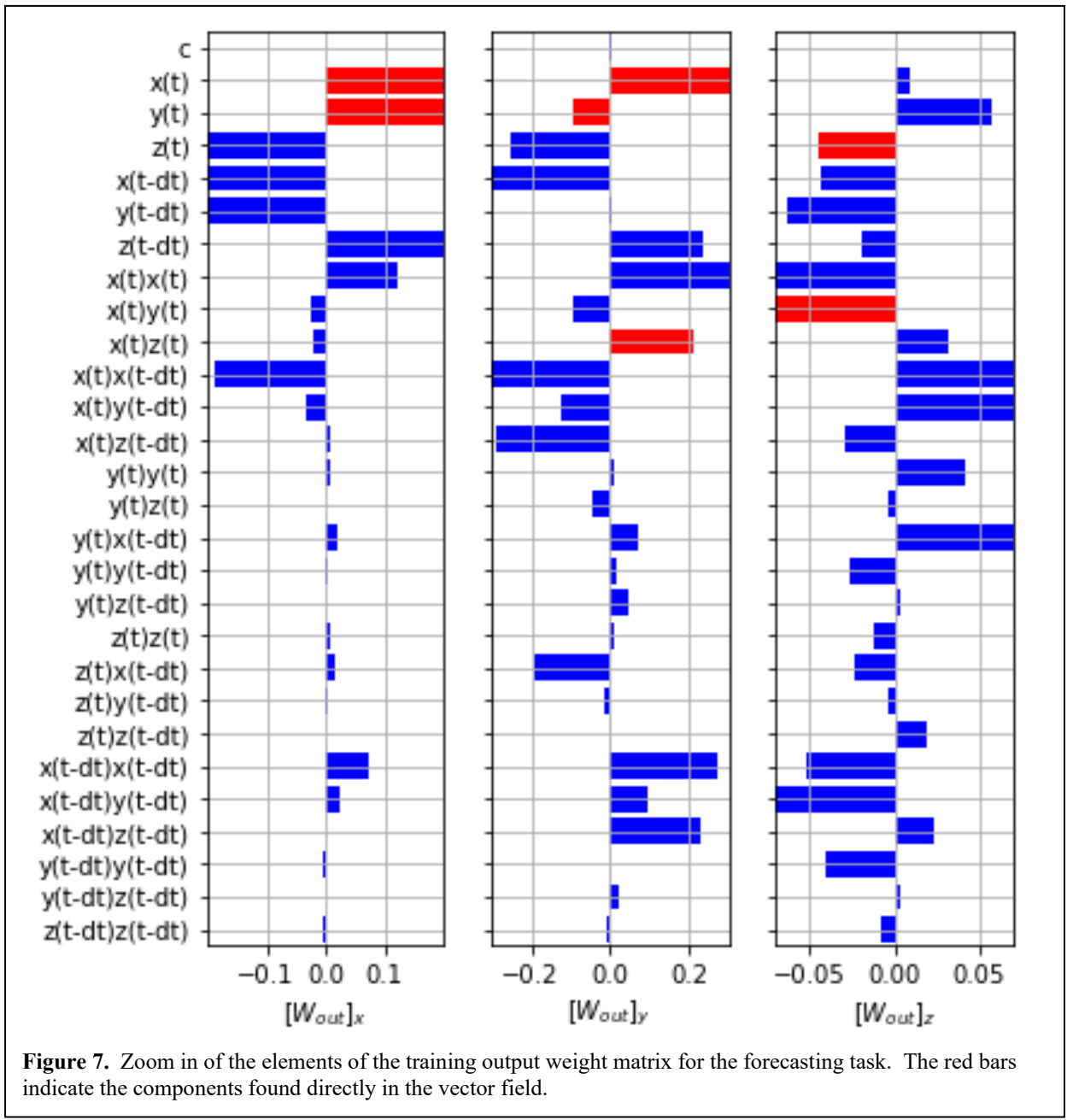

**Figure 7.** Zoom in of the elements of the training output weight matrix for the forecasting task. The red bars indicate the components found directly in the vector field.

*Inference Task*

Figure 8 shows the components of the output weight matrix for the inference task presented in Fig. 4, where the $z$ is inferred given $x$ and $y$. The largest component of $\mathbf{W}_{out}$ is the constant term $c$. It can be explained by the fact that $z$ has a considerable offset as shown in the time series of Fig. 2 of the main text, although all the other $\mathbf{W}_{out}$ components are non-zero and contribute for the final output to some extent.

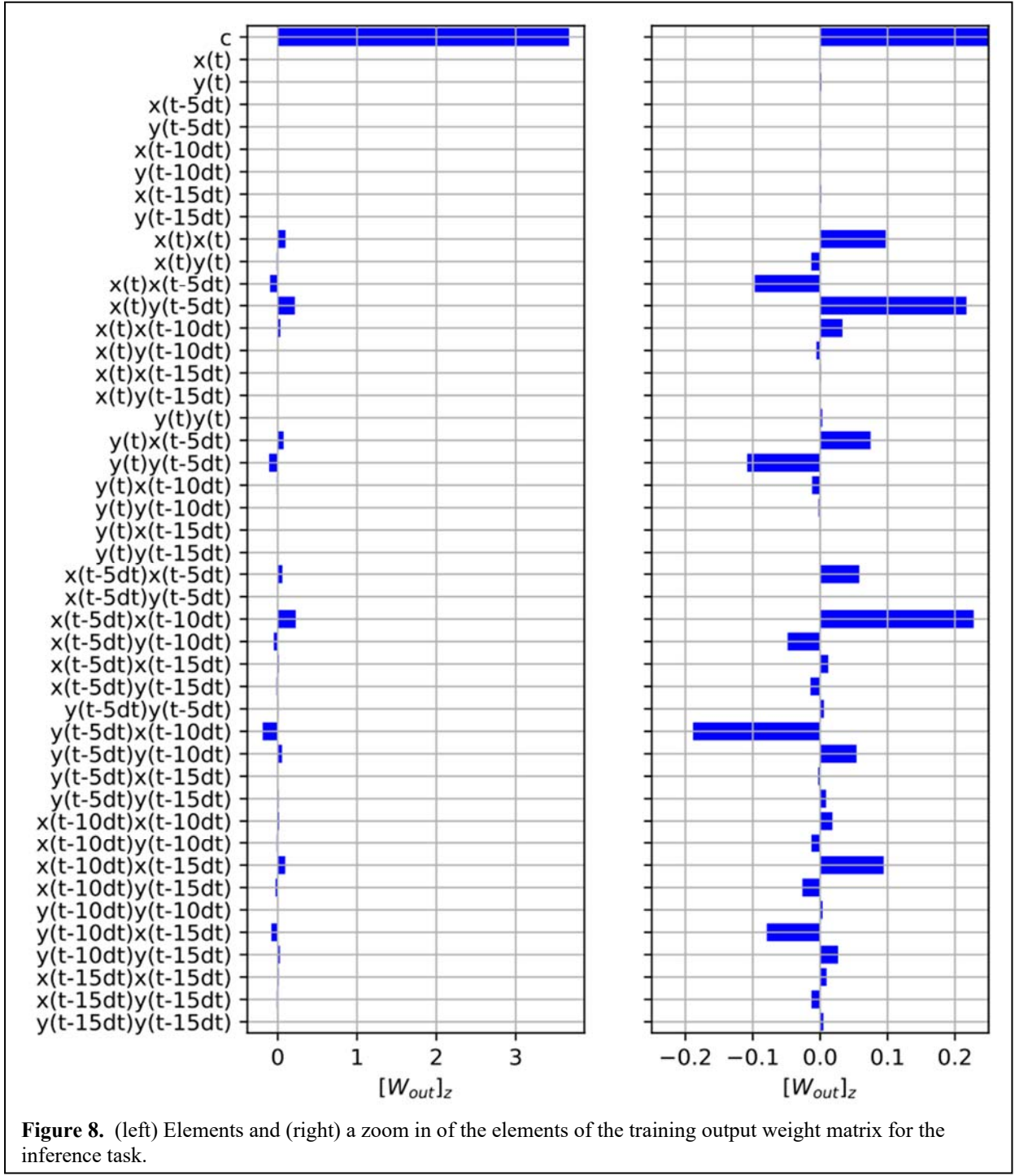

**Figure 8.** (left) Elements and (right) a zoom in of the elements of the training output weight matrix for the inference task.

*Comparing the computational complexity of the NG-RC with a typical traditional RC*

Here, we provide an indication of computational speed up for the NG-RC compared to a traditional RC for the Lorenz63[3,9,10] (quartic nonlinear output layer for the NG-RC) and double-scroll[9] (cubic nonlinear output layer for the NG-RC) forecasting tasks by estimating the number of multiplications and special function evaluations for each. Our assumptions and notation are as follows, where we only track parameters that contribute the most to the computational complexity.

NG-RC and traditional RC parameters:
  Warm-up steps: $M_{warmup}$
  Training steps: $M_{train}$

NG-RC-specific parameters:
  Dimension of the linear part of the feature vector: $N_{linear}$
  Multiplications need to form feature vector: $N_{nonlinear}$
  Total components in the feature vector: $N_{total}$

Traditional RC-specific parameters:
  Number of reservoir nodes: $N$
  Sparsity of the internal reservoir connections: $\sigma_r$
  Number of special function evaluations (typically tanh): $N_{special}$

For the traditional RC, we assume sparse connectivity and that there is no additional overhead for using sparse matrix multiplication routines. Special function evaluations can be computationally expensive or not depending on built-in mathematical co-processors. We give the number of these evaluations but do not use it in comparing the computational complexity.

The dominant contribution to the computational complexity for the NG-RC is performing the ridge regression, which is $\mathcal{O}((M_{train}(N_{total})^2)$ over the training time. The dominant contribution for the traditional RC is multiplying the reservoir state with the adjacency matrix, which is $\mathcal{O}(\sigma_r (M_{warmup}+M_{train}) (N_{total})^2)$ over the warmup and training time, and performing the ridge regression, which is $\mathcal{O}(M_{train}(N_{total})^2)$ over the training time assuming all nodes contribute to the prediction. Also, we have $N_{special} = N$, but we do not consider this computational cost.

We estimate the speed up by summing the dominant contributions for the traditional RC and divide by the sum for the NG-RC. The traditional RC used in Ref. 9 is meant to be efficient (fast simulation time) at the expense of some accuracy, the one used in Ref. 10 is meant to have high accuracy at the expense of longer simulation time, while the one used in Ref. 3 is intermediary. Clearly, our analysis indicates a substantial speed up even with our conservative analysis, while the NG-RC simultaneously obtains high accuracy.

**Table 1: Estimated speed up of the NG-RC for the Lorenz63 forecasting task**

|  | $M_{warmup}$ | $M_{train}$ | $N_{nonlinear}$ | $N_{total}$ | $N$ | $\sigma_r$ | speed up |
|---|---|---|---|---|---|---|---|
| NG-RC | 2 | 400 | 21 | 28 | - | - | - |
| Ref. 9 | 1,000 | 1,000 | - | 100 | 100 | 0.01-0.05 | 33-163 |
| Ref. 3 | ? (set to 0) | 5,000 | - | 300 | 300 | 0.02 | $1.5 \times 10^3$ |
| Ref. 10 | $10^5$ | $6 \times 10^4$ |  | 4,000 | 2,000 | 0.02 | $3.2 \times 10^6$ |

**Table 2: Estimated speed up of the NG-RC for the double-scroll forecasting task**

|  | $M_{warmup}$ | $M_{train}$ | $N_{nonlinear}$ | $N_{total}$ | $N$ | $\sigma_r$ | speed up |
|---|---|---|---|---|---|---|---|
| NG-RC | 2 | 400 | 56 | 62 | - | - | - |
| Ref. 9 | 1,000 | 1,000 | - | - | 100 | 0.01-0.05 | 8-41 |

*NG-RC performance with training data set size*

We do not yet have an analytic expression for predicting the required training data set size for an RC. We hypothesize that a lower bound on the training data set size is about the number of unknown fit parameters, equal to the number of features times the dimension of the forecasted system (3 in our examples). For the NG-RC and the Lorenz63 task, this is 84 (see Table 1), whereas it is 12,000 for Ref. 10. This is approximately the minimum number of data points needed so that the model passes exactly through the training data points. To generalize the model to unseen data, as required for forecasting, requires some additional data overhead.

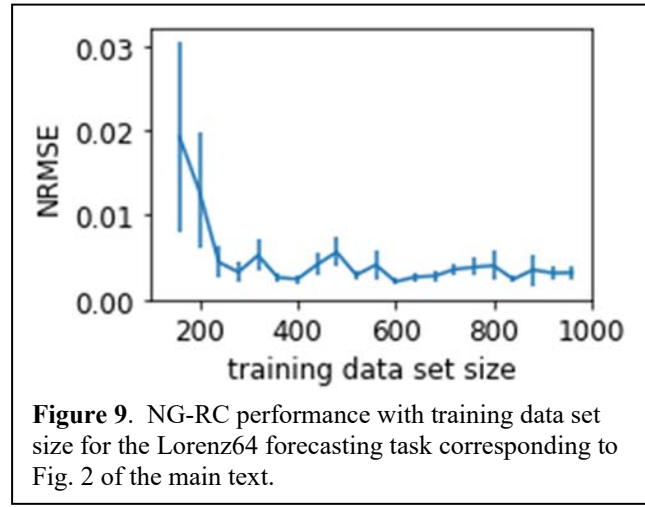

**Figure 9**. NG-RC performance with training data set size for the Lorenz64 forecasting task corresponding to Fig. 2 of the main text.

Figure 9 shows the training data set size dependence of the testing NRMSE, averaged over 20 different temporal segments, for $\alpha = 2.5 \times 10^{-6}$ for all cases. For low training data, there are large fluctuations in the error and there is greater sensitivity to changes in $\alpha$. Around 250 training data points, the error saturates, indicating that no more data is needed for good, generalized performance. The NRMSE is only one measure of performance; we find that the return map, discussed above, is more sensitive to the training data set size and we find small, improved performance beyond the 400-point set used in the main text.

We stress that the training data set size required for good performance also depends on the sampling step size *dt*. If dt is too small, only a small region of the attractor is visited for a small number of data points. If *dt* is too large, higher-order nonlinear features might be needed. Our choice of *dt*, which gives about 40 points per Lyapunov time, balances these two constraints and leading to a small, required data set.

*NG-RC performance for the Lorenz63 system driven by noise*

Here, we explore the ability of the NG-RC when a dynamical system is driven by large noise. Here, we augment the differential equations for the Lorenz63 system by adding Gaussian random noise to the right-hand-side of the differential equations for each variable with a root-mean-square value of 1. This noise level is about 12% of the typical root-mean-square values of each variable, which are equal to 7.9, 9.0, and 8.6 for *x*, *y*, and *z*, respectively.

To obtain good generalization, we find that we need to increase the ridge parameter to $\alpha = 1.4 \times 10^{-2}$. The Lorenz63 dynamics and NG-RC fit are shown in Figs. 9a)-d).

After the model is trained, we use the NG-RC to forecast the Lorenz63 dynamics initialized by the last point in the training data set. We then compare the predicted behavior to the *noise-free* Lorenz63 dynamics and find that the RMSE = $1.34 \times 10^{-2}$. This low error indicates that the NG-RC learns the underlying deterministic system even when it is perturbed by substantial noise. We find similar behavior when measurement noise is added to the time-series data (not shown).

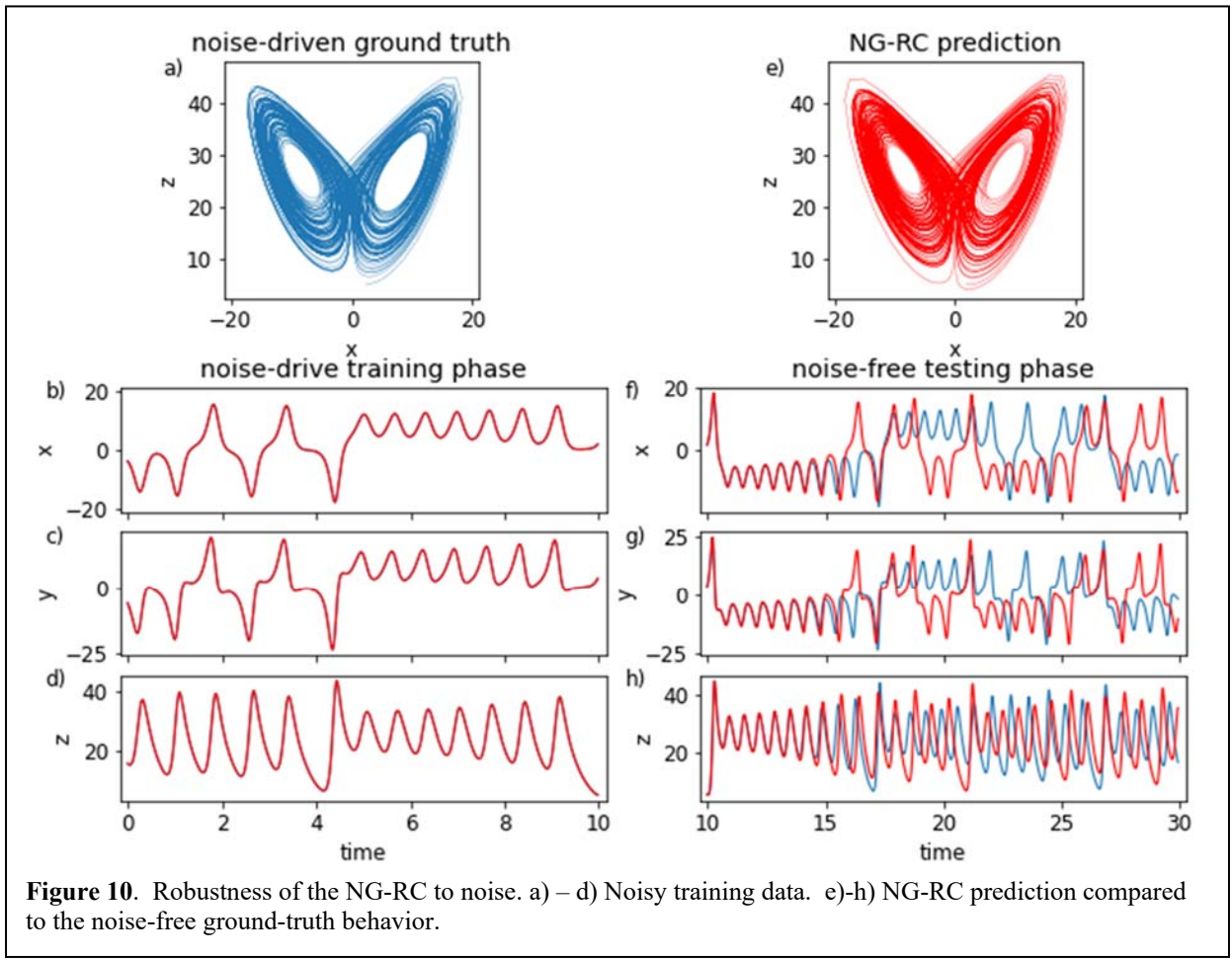

**Figure 10**. Robustness of the NG-RC to noise. a) – d) Noisy training data. e)-h) NG-RC prediction compared to the noise-free ground-truth behavior.